%% file: root.tex
\documentclass[letterpaper, 10 pt, conference]{ieeeconf}

\IEEEoverridecommandlockouts
\pdfoutput=1 
\usepackage{xspace}
\usepackage{amssymb,paralist,epsfig,standalone,bm,placeins}  
\usepackage{graphicx} 
\usepackage{xcolor}
\usepackage[utf8]{inputenc}
\usepackage{amsmath,amsfonts,booktabs,cite} 
\usepackage{siunitx,textcomp} 
\usepackage[hidelinks]{hyperref} 
\let\labelindent\relax
\usepackage[inline]{enumitem} 
\usepackage[acronym]{glossaries}

\usepackage{tikz,standalone,pgfplots}
\usepackage{subfig}
\usetikzlibrary{patterns}
\usepackage{pgfplots, pgfplotstable}
\usepackage{caption}
\graphicspath{{figures_tex/}}

\usepackage{changes}

\pgfplotsset{compat=1.12}

\input{acronyms.tex}

\input{macro.tex}

\definecolor{findOptimalPartition}{HTML}{D7191C}
\definecolor{storeClusterComponent}{HTML}{FDAE61}
\definecolor{dbscan}{HTML}{ABDDA4}
\definecolor{constructCluster}{HTML}{2B83BA}

\title{\LARGE \bf
A Novel Simulation-Based Quality Metric for Evaluating Grasps on\\ 3D Deformable Objects
}

\author{Tran~Nguyen~Le, Jens~Lundell, Fares~J.~Abu-Dakka, Ville~Kyrki%
\thanks{This work was supported by Academy of Finland Strategic Research Council
grant 314180 and CHIST-ERA project IPALM (326304). We gratefully acknowledge the support of
NVIDIA Corporation with the donation of the Titan Xp GPU used for this research.} \thanks{All authors are with
Intelligent Robotics Group at the Department of Electrical Engineering and
Automation, School of Electrical Engineering, Aalto University, Finland.
\texttt{\{firstname.lastname\}{@}aalto.fi}}
}

\begin{document}
\maketitle
\thispagestyle{empty}
\pagestyle{empty}


\begin{abstract}

Evaluation of grasps on deformable 3D objects is a little-studied problem, even if the applicability of rigid object grasp quality measures for deformable ones is an open question. 
A central issue with most quality measures is their dependence on contact points which for deformable objects depend on the deformations. This paper proposes a grasp quality measure for deformable objects that uses information about object deformation to calculate the grasp quality. Grasps are evaluated by simulating the deformations during grasping and predicting the contacts between the gripper and the grasped object. The contact information is then used as input for a new grasp quality metric to quantify the grasp quality. The approach is benchmarked against two classical rigid-body quality metrics on over 600 grasps in the Isaac gym simulation and over 50 real-world grasps. Experimental results show an average improvement of 18\% in the grasp success rate for deformable objects compared to the classical rigid-body quality metrics.   

\end{abstract}

\input{sections/introduction}

\input{sections/related_work}
\input{sections/grasp_evaluation}

\input{sections/experiment}

\input{sections/conclusion}

\bibliographystyle{IEEEtran}
\bibliography{refs}

\end{document}

%% file: acronyms.tex
\newacronym{mss}{MSS}{Mass-Spring System}
\newacronym{pbd}{PBD}{Position-based Dynamics}
\newacronym{fem}{FEM}{Finite Element Method}
\newacronym{dnn}{DNN}{Deep Neural Network}
\newacronym{fcn}{FCN}{fully-convolutional network}
\newacronym{gws}{GWS}{grasp wrench space}
\newacronym{gqm}{GQM}{Grasp Quality Metric}

%% file: macro.tex
\newcommand{\figref}[1]{\hyperref[#1]{Fig.~\ref*{#1}}}
\newcommand{\tabref}[1]{\hyperref[#1]{Table~\ref*{#1}}}
\newcommand{\secref}[1]{\hyperref[#1]{Section~\ref*{#1}}}
\newcommand{\algoref}[1]{\hyperref[#1]{Algorithm~\ref*{#1}}}

\newcommand{\ra}[1]{\renewcommand{\arraystretch}{#1}}
\newcommand{\tbs}[1]{\renewcommand{\tabcolsep}{#1pt}}

\def\shake{\textit{shake task}\xspace}

\def\panda{Franka Emika Panda\xspace}

\def\etal{\textit{et al.} }

\def\figvspace{\vspace{-1.2em}}

%% file: sections/introduction.tex
\section{Introduction}
\label{sec:introduction}

\begin{figure}[!t]
    \centering
	\includegraphics[width=\linewidth]{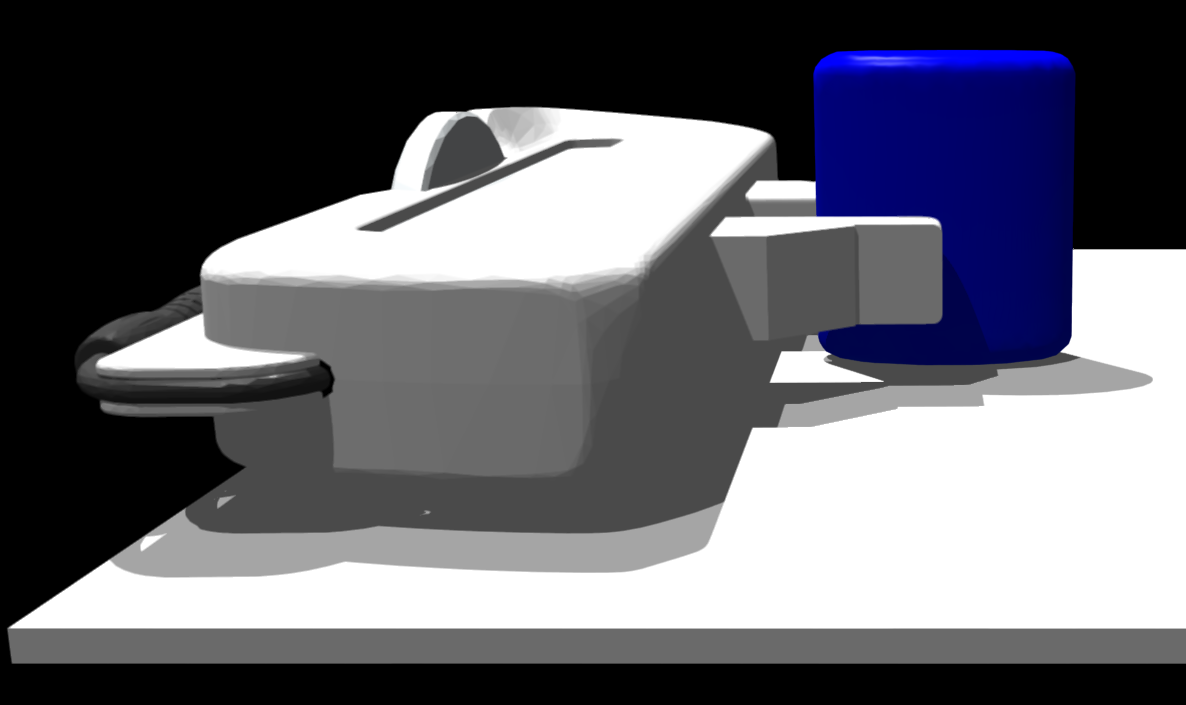}  \\
	\includegraphics[width=.4\linewidth]{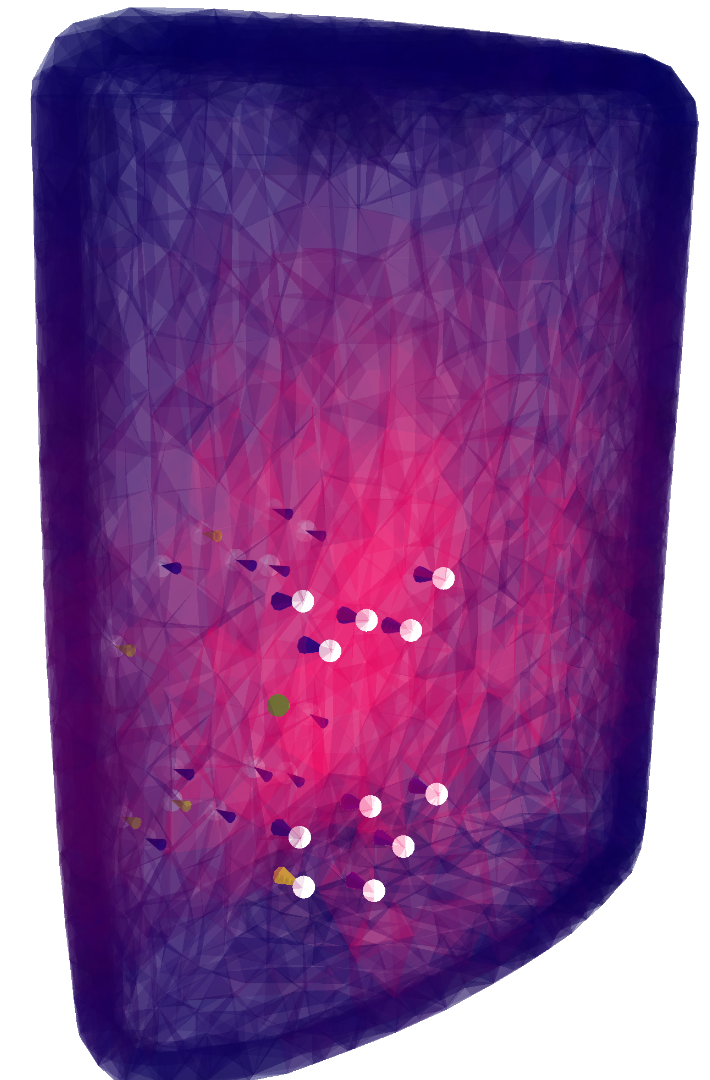}  
    \hspace{.5 em}
	\includegraphics[width=.4\linewidth]{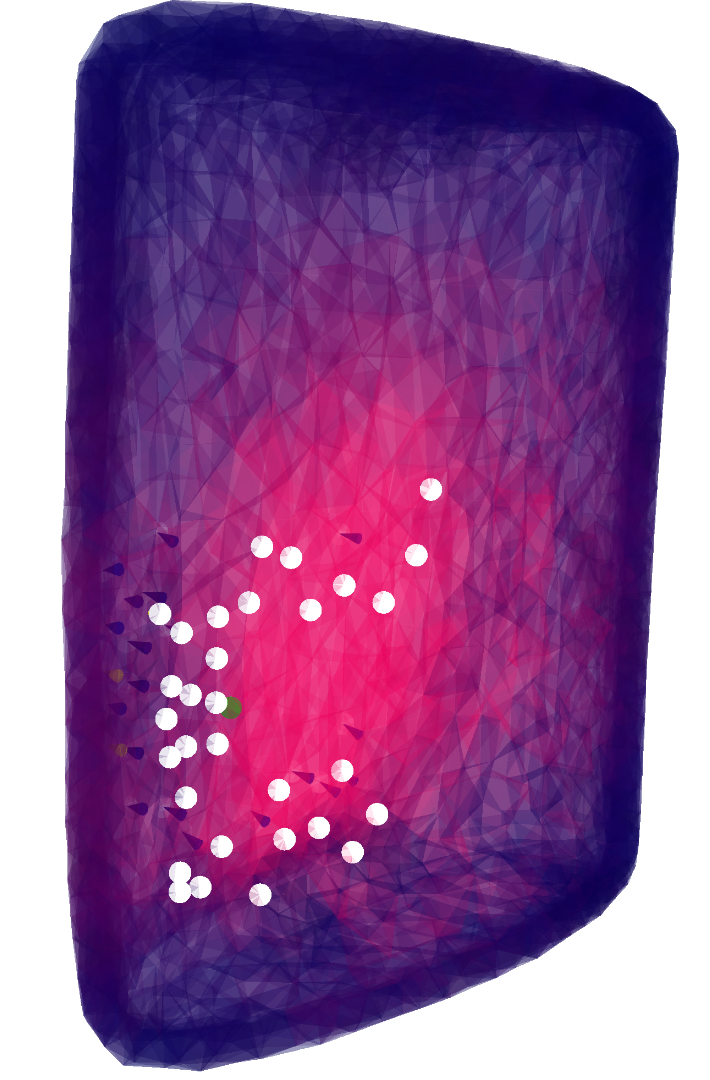}
    \caption{The top-image shows an example grasps on a deformable object executed in the Isaac Gym simulator. The bottom images show the amount of deformation the object undergoes (the redder the object the more it deforms) for a 2N grasping force (bottom-left) and a 5N grasping force (bottom-right). With more object deformation, more contact points displayed as white spheres form between the gripper and the object. 
    }
    \label{fig:title}
    \vspace{-1.5em}
\end{figure}

Consider the parallel-jaw grasp in \figref{fig:title}. Is that a good or a bad grasp? To answer such a question one often resorts to calculating a \gls{gqm}, which is an index that quantifies the goodness of a grasp. To date, many different \glspl{gqm} exist, including the Ferrari-Canny epsilon quality metric \cite{ferraricanny}, the volume quality metric \cite{volumeofgws}, and many more \cite{graspqualityreview}. What most of these quality metrics have in common, though, is the assumption that the object is rigid. What follows from such an assumption is that contacts between the finger and the object can be treated as static single point contacts, which simplifies the calculation of the \glspl{gqm}. However, as shown in \figref{fig:title}, when grasping a non-rigid object, the deformation of the object due to the grasping force will continuously form multiple contact points per finger, forcing us to rethink how to do grasp evaluation on such objects.

Only recently have some researchers started to investigate how to evaluate grasps on deformable objects \cite{defgraspsim}. The approach proposed in \cite{defgraspsim} evaluated grasps on deformable objects by simulating a whole dynamic manipulation task after the robot grasped the object. Although a reasonable approach, it was computationally demanding with running times in the order of minutes. In this paper, instead of simulating a whole dynamic manipulation task, we present a computationally cheaper analytical approach for evaluating grasps on deformable objects. Our approach consists of a \gls{fem}-based simulation of grasps to continuously model changes in shape and contact geometry between the gripper and the object. Based on the \gls{fem}-based simulation, we can continuously record contact specific data such as contact locations, normals, and wrenches which are used to calculate a time-dependent \gls{gws}. Using the time-dependent \gls{gws}, we also propose a novel gravity-resistant \gls{gqm} that quantifies grasps according to the direction and magnitude of the smallest gravitational force that would break a grasp. 

We experimentally evaluate the proposed approach and quality metric on over 600 grasps in the Isaac Gym simulator and on 50 real-world grasps using a Franka Emika Panda robot. All experimental results suggest that the proposed approach, together with the new \gls{gqm}, is more suitable for grasp evaluation on deformable objects than classical rigid-body \glspl{gqm}.   

In summary, the main contributions of this paper are:
\begin{itemize}
    \item A \gls{fem}-based simulation approach for evaluating grasps on deformable objects.
    \item A novel gravity-resistant \gls{gqm} that considers both grasp wrench resistance and object deformation during the grasp.
    \item A thorough empirical evaluation of the proposed approach and quality metric presenting, both in simulation and on real hardware, improvements over existing quality metrics in terms of grasp quality evaluation and grasp success rates.
\end{itemize}

%% file: sections/related_work.tex
\section{Related work}
\label{sec:related_work}
The work presented in this paper addressed grasp evaluation on deformable objects and soft-body contact simulation. Therefore, we will review both of these areas separately. In addition, we will review works on grasp evaluation on rigid objects as it is central to the approach we develop in this paper.

\subsection{Grasp Evaluation on Rigid Objects}
The overarching goal of grasp evaluation is to quantify the goodness of a grasp using mathematical models of the interaction between the manipulator and the object. One main reason for quantifying the goodness of grasps is that it enables us to search for better grasps \cite{lin15,mahler2017dex,mousavian2019graspnet}.  

To quantify a grasp, we often resort to calculating a \gls{gqm}. Throughout the grasping history, many different \glspl{gqm} have been proposed \cite{graspqualityreview}, and among these the most popular and well-established metrics are most likely the Ferrari-Canny epsilon metric \cite{ferraricanny} and the volume quality metric \cite{volumeofgws}. What the epsilon, volume, and many of the other quality metrics have in common is that they assume the object to grasp is rigid. 

However, recent works \cite{deformmechanics,yan14,defgraspsim,ken_deform,wakamatsu_bfc} have demonstrated that the aforementioned metrics are unsuitable for and cannot be directly applied to evaluating grasps on deformable objects due to object deformation under interaction forces. Therefore, we propose a new quality metric tailored for evaluating grasps on deformable objects in this work. Additionally, we compare our new quality metric to the epsilon and volume quality metrics to determine which metric is most suitable for evaluating grasps on deformable objects.  

\subsection{Grasp Evaluation on Deformable Objects}
Grasp evaluation on deformable objects is, compared to rigid objects, a less explored research area. Only recently have a few works proposed quality metrics or methods to evaluate the  goodness of a grasp on deformable objects \cite{wakamatsu_bfc,alt_minimize,xu_minimalwork,ken_deform,defgraspsim,peng22}. 

One of the above mention works \cite{xu_minimalwork}, target the problem of minimizing the deformation of 3D hollow deformable objects. To achieve minimal deformation, the authors defined the \textit{minimal work} \gls{gqm}. The minimal work \gls{gqm} differs from the quality metrics proposed for rigid objects by including a local stiffness map of the object as a component to the wrench-space. The local stiffness map was estimated by squeezing the object in simulation \cite{alt_minimize} or using real physical hardware. The main limitation of the approach in \cite{xu_minimalwork} is that it only works on objects with small deformations.

In contrast to minimizing deformation, some works have proposed to utilize object deformation to successfully accomplish tasks \cite{wakamatsu_bfc,ken_deform,defgraspsim,peng22}. The earliest of these was \cite{wakamatsu_bfc}, where the authors introduced the concept of bounded force closure, which extends the concept of force closure from rigid objects to deformable objects. In a nutshell, bounded force closure can determine if a grasp on a deformable object is force closure given a maximal allowable external force. Similarly, Gopalakrishnan and Golberg~\cite{ken_deform} introduced \textit{deform closure}, which is a generalization of the form closure concept on rigid objects to deformable objects with frictionless contact. In essence, a grasp is defined as deform closure when positive work is needed to release the object at its deformed configuration. 

More recently, Isabella~\etal~\cite{defgraspsim} have proposed to use a \gls{fem}-based soft-body simulator to run a battery of different grasp performance metrics for evaluating grasps on deformable objects. They also show that these performance metrics are indicative of real-world grasp success. Nevertheless, the main limitation of the approach proposed in \cite{defgraspsim} is the high computational cost for simulating a grasp, which often requires several minutes or more. A similar approach to \cite{defgraspsim} was recently presented in \cite{peng22}. In \cite{peng22}, the authors dynamically evaluated grasps using dynamic grasp maps and existing \glspl{gqm}. However, the author only validated the metrics on one object and four different grasp candidates, and as such, we cannot draw extensive conclusions from the results.

In this work, we take inspiration from \cite{defgraspsim,peng22}, and leverage the idea of continuously modelling the contact between the robot and the deformable objects during a grasp and using this information for calculating the quality of a grasp. However, instead of using grasp performance metrics as in \cite{defgraspsim} or already define quality metrics as in \cite{peng22}, we propose a new analytical quality metric that is specifically tailored for 3D deformable objects. 

\subsection{Soft-body Simulation}
\label{sec:softsim}
In terms of soft-body simulation, the choice of geometric representation heavily affects the behavior of simulating dynamics of deformable objects. In a recent survey on deformable object manipulation \cite{yin_survey}, Yin~\etal presented three primary deformable object modelling approaches: \gls{mss}, \gls{pbd}, and \gls{fem}. Among these three, \gls{fem} offers a more physically accurate representation of a deformable object despite its computational cost. In addition, \gls{fem}  performs well on objects with large deformations. Therefore, in this work, we propose using \gls{fem} to model and simulate the contact between the object and the robot gripper. 

In terms of actual simulation, many different physics simulators exist, including MuJoCo \cite{mujoco}, PyBullet \cite{pybullet}, SOFA\cite{sofa}, and the more recent Isaac Gym simulator \cite{isaacgym_ofi}. Out of these, SOFA and Isaac Gym are primary simulators for soft bodies that support \gls{fem}. In addition, Isaac Gym has a grasping framework named DefGraspSim \cite{defgraspsim} to automatically perform and evaluate grasp tests on  deformable and rigid objects. Because of these benefits, we chose Isaac Gym to simulate soft-bodies and DefGraspSim to evaluate our proposed approach.


%% file: sections/grasp_evaluation.tex
\section{Grasp Evaluation on 3D Deformable Objects}
\label{sec:prob}

This paper addresses the problem of evaluating parallel-jaw grasps on 3D deformable objects lying on a supporting surface. This implies producing a \gls{gqm} $\mathcal{Q}$ that measures how well a grasp can keep the object restrained even if affected by disturbing forces or torques.

We assume the geometry and stiffness of the target object are known and that the object's origin is at its center-of-mass. We also consider that the direction of gravity is unknown.
\subsection{Background}
\label{sec:background}
Let us consider a grasp with $N$ contact points. Through each of these contact points, a robot can transfer wrenches $\bm{w}_n \in \mathbb{R}^6$ to the object to counteract disturbances. A wrench is a six-dimensional vector consisting of a force $\bm{f}_n \in \mathbb{R}^3$ and a torque component $\bm{\mathcal{T}}_n \in \mathbb{R}^3$. 
If we consider the frictional contact point model, the contact wrench at location $\bm{x}_n \in \mathbb{R}^3$ is defined as:

\begin{equation*}
     \bm{w}_n = \left[\begin{array}{c}\bm{f}_n\\\bm{\mathcal{T}}_n\end{array}\right] = \left[\begin{array}{c}\bm{f}_n\\\bm{x}_n \times \bm{f}_n\end{array}\right].
\end{equation*}

Given all of the wrenches $\bm{w}_n~\forall n,\dots,N$, we construct the \gls{gws} $\mathcal{P}$ as 
\begin{equation*}
     \mathcal{P} = \text{ConvexHull}\left(\bigcup_{n=1}^{N}\bm{w}_n\right).
\end{equation*}
 The \gls{gws} is at the center of many different grasp quality metrics, including the well-known Ferrari-Canny epsilon quality metric \cite{ferraricanny} and the volume quality metric \cite{volumeofgws}, both of which we will use in the experimental evaluation in \secref{sec:exp_and_res}. In detail, the epsilon quality metric describes the minimum wrench needed to break a grasp and is defined as the radius of the largest 6D sphere centered at the origin enclosed by the convex hull of the wrench space. The volume metric, on the other hand, describes the average force a grasp can withstand and is defined as the volume of the wrench space’s convex hull.

The main limitation of the above quality metrics is that they assume rigid point contacts between the gripper and the object. In other words, no matter the force executed by the gripper, the contact points will always be the same as the initial ones. This assumption is valid for rigid objects that do not deform under contact. However, for non-rigid objects, the assumption breaks down as the contact points continuously change during the execution of the grasp or the disturbances due to object deformations. Therefore, to enable evaluating grasps on 3D deformable objects, we propose an approach that consists of a time-dependent \gls{gws} and a gravity-resistant quality metric.

\subsection{Time-Dependent \gls{gws}}

As non-rigid objects deform under interaction forces, the initial contact locations will change, new contacts might form, and old contact locations might disappear. Because of this fact, grasps on deformable objects need to be defined as temporally evolving multi-contact points. 

To allow for temporally evolving multi-contact points, we need to track contact points. For this purpose, we propose using the \gls{fem} to gather necessary contact information for calculating the \gls{gws}, such as contact locations, the contact normals, and contact wrenches. Based on this contact information and using the same frictional contact point model as before, we redefine the \gls{gws} to be time-dependent:
\begin{equation*}
     \mathcal{P}(t) = \text{ConvexHull}\left(\bigcup_{n=1}^{N}\bm{w}_n(t)\right).
\end{equation*}

The time-dependent \gls{gws} enables computing a \gls{gqm} at each time step. It is important to note that in this work instead of using the typical \gls{gws} which is usually centered at the object center-of-mass, we consider \textit{contact-centered} \gls{gws} where we center \gls{gws} at the center of contact points. The reason for this choice is that we want to study not only the behavior of the interaction between the gripper and the target object but also the effect of the disturbances on the grasp at the contact regions. In the next section, we will detail a new specific quality metric that uses the time-dependent \gls{gws} for evaluating grasps on deformable objects.

\subsection{Gravity-Resistant Quality Metric}
\begin{figure}
    \centering
    \def\svgwidth{\linewidth}
     {\color{black} \fontsize{15}{15}
    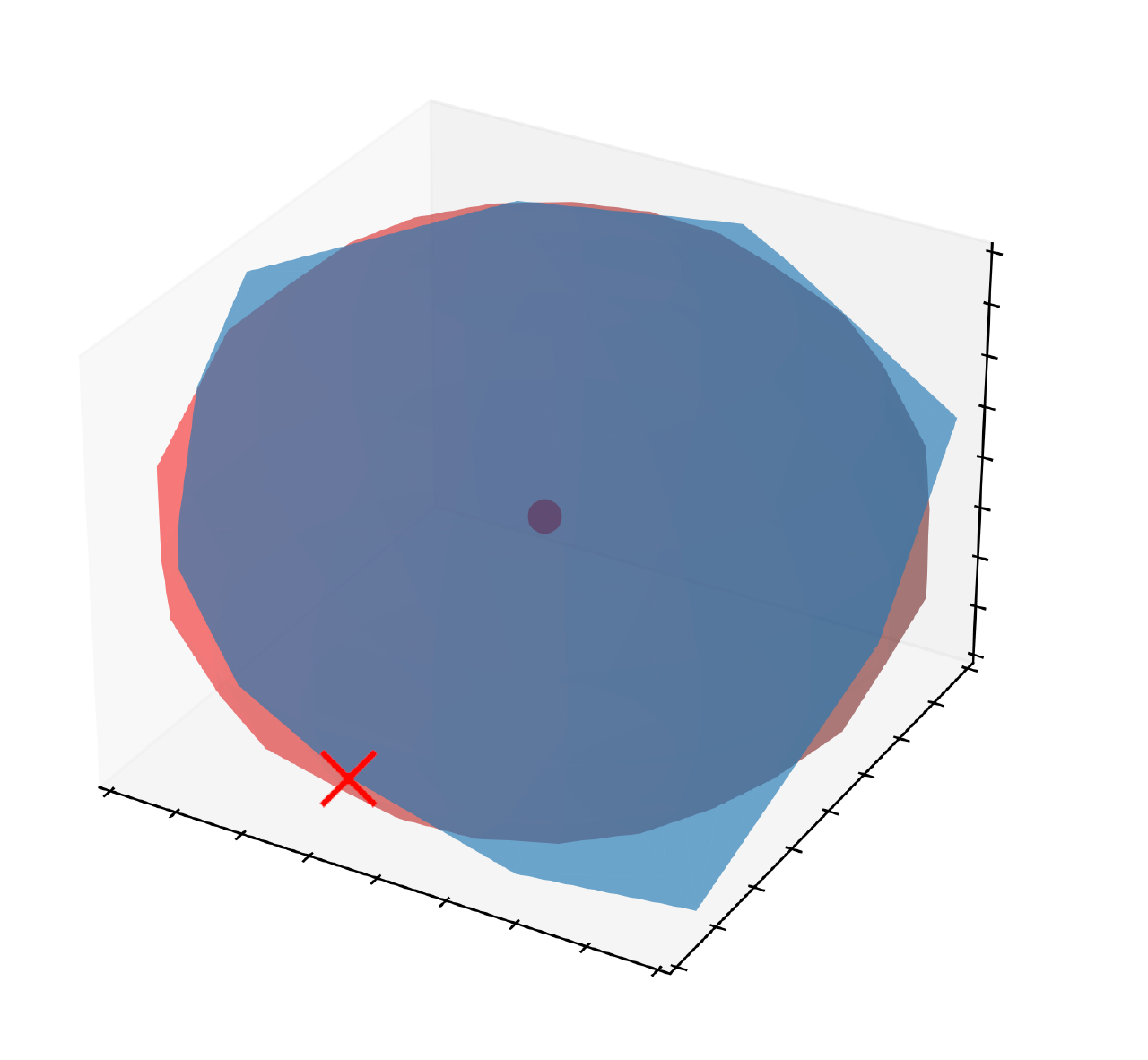}       
    \caption{$\mathcal{P}(t)$ (blue polygon), $\mathcal{P}_{gravity}$ (red sphere), $\mathcal{P}_{int}$ (green dashed polygon) projected into force space. The proposed quality metric is then determined as the smallest distance between the origin of the wrench spaces visualized as the red dot and the closest intersection point of the two wrench spaces visualized as the red cross. The larger the distance, the higher the quality metric and the better the grasp.}
    \label{fig:intersect}
    \vspace{-0.5cm}
\end{figure}

Based on previous work \cite{defgraspsim,lin_feel3d,com16}, the main disturbance that causes grasp failure on deformable objects is gravity. Therefore, we propose a quality metric that can quantify how well a grasp can safely keep the object under the effect of gravity. Our quality metric is inspired from \cite{inspire14} and based on the mechanical wrench space analysis outlined in \secref{sec:background}. However, the \gls{gqm} in \cite{inspire14} only consider the stability of soft-finger grasp on rigid objects under gravity and not the stability of rigid grasps on non-rigid objects which is the problem we target.

As we want to counteract the gravity induced disturbances, we first define the gravitational 6D wrench space acting on the object at object center-of-mass as $\mathcal{P}_{gravity}$. However, as we want to evaluate a grasp in the case where the gravitational force direction is unknown, we need to define an unknown directional gravitational wrench. We approximate such a wrench by calculating the \gls{gws} $\mathcal{P}_{gravity}$ as if the wrench was acting on the object from 16 regularly spaced directions sampled from the unit sphere. A visual example of $\mathcal{P}_{gravity}$ is shown in \figref{fig:intersect}. 

Next, we calculate the intersection of $\mathcal{P}(t)$ and the unknown directional gravitational \gls{gws} $ \mathcal{P}_{gravity}$ as
\begin{equation}
    \label{eq:int_P}
    \mathcal{P}_{int}(t) = \mathcal{P}(t) \cap \mathcal{P}_{gravity}.
\end{equation}

Given the intersected wrench space calculated using \eqref{eq:int_P}, we define our new quality metric as the smallest distance between the origin and the border of this wrench space
\begin{equation*}
    \mathcal{Q}(t) = \min_{\bm{w} \in \mathcal{P}_{int}(t)} || \bm{w} ||.
\end{equation*}
This metric indicates in the direction and the magnitude of the smallest gravitational force the grasp is able to withstand before breaking. \figref{fig:intersect} shows an example of the two different \glspl{gws}, the intersection wrench space, and the proposed quality. 

By continuously determining the \gls{gws} $\mathcal{P}(t)$ for different time-steps, we can evaluate the quality of a grasp candidate throughout the grasp, even as objects deform. In the next section, we will experimentally validate if this leads to more successful grasps.



%% file: figures_tex/intersectnew.pdf_tex
\begingroup%
  \makeatletter%
  \providecommand\color[2][]{%
    \errmessage{(Inkscape) Color is used for the text in Inkscape, but the package 'color.sty' is not loaded}%
    \renewcommand\color[2][]{}%
  }%
  \providecommand\transparent[1]{%
    \errmessage{(Inkscape) Transparency is used (non-zero) for the text in Inkscape, but the package 'transparent.sty' is not loaded}%
    \renewcommand\transparent[1]{}%
  }%
  \providecommand\rotatebox[2]{#2}%
  \newcommand*\fsize{\dimexpr\f@size pt\relax}%
  \newcommand*\lineheight[1]{\fontsize{\fsize}{#1\fsize}\selectfont}%
  \ifx\svgwidth\undefined%
    \setlength{\unitlength}{368.63999176bp}%
    \ifx\svgscale\undefined%
      \relax%
    \else%
      \setlength{\unitlength}{\unitlength * \real{\svgscale}}%
    \fi%
  \else%
    \setlength{\unitlength}{\svgwidth}%
  \fi%
  \global\let\svgwidth\undefined%
  \global\let\svgscale\undefined%
  \makeatother%
  \begin{picture}(1,0.92317712)%
    \lineheight{1}%
    \setlength\tabcolsep{0pt}%
    \put(0,0){\includegraphics[width=\unitlength,page=1]{intersectnew.pdf}}%
    \put(0.16034347,0.15214268){\rotatebox{-19.258436}{\makebox(0,0)[lt]{\lineheight{1.25}\smash{\begin{tabular}[t]{l}$f_{x}$\end{tabular}}}}}%
    \put(0.71995146,0.12737814){\rotatebox{48.293286}{\makebox(0,0)[lt]{\lineheight{1.25}\smash{\begin{tabular}[t]{l}$f_{y}$\end{tabular}}}}}%
    \put(0.90931353,0.5009106){\rotatebox{83.997348}{\makebox(0,0)[lt]{\lineheight{1.25}\smash{\begin{tabular}[t]{l}$f_{z}$\end{tabular}}}}}%
    \put(0,0){\includegraphics[width=\unitlength,page=2]{intersectnew.pdf}}%
  \end{picture}%
\endgroup%

%% file: sections/experiment.tex
\section{{Experiments and Results}}
\label{sec:exp_and_res}
The two main questions we wanted to answer with the experiments were:
\begin{enumerate}
    \item Is the proposed grasp quality metric better than rigid-object quality metrics at finding successful grasp candidates on deformable objects?    
    \item Do the high-quality grasps proposed by our quality-metric succeed in picking up objects in the real world?
\end{enumerate}
To answer the first question we performed experiments in simulation and to answer the second question we performed real-world experiments.

\subsection{Grasp Quality Evaluation in Simulation}
\begin{figure}[t]
	\centering
    \def\svgwidth{\linewidth}
     {\color{black} \fontsize{15}{15}
    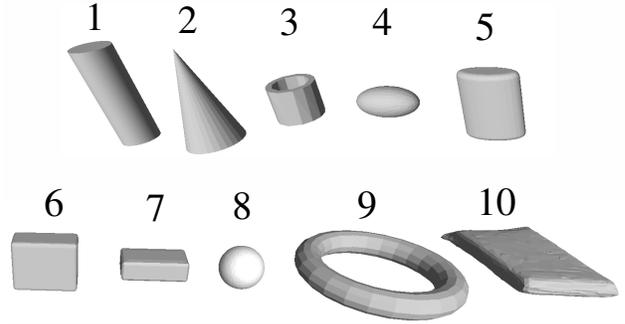}   
	\caption{The ten deformable objects used in the simulation experiment.}
	\label{fig:simobjects}
	\vspace{-1em}
\end{figure}

To investigate the proposed grasp quality metric in simulation, we used the Isaac Gym simulator with the DefGraspSim framework  \cite{defgraspsim} on the ten deformable objects shown in \figref{fig:simobjects}. Each object is represented as a tetrahedral mesh placed on a rectangular platform. The stiffness of an object is changed by adjusting Young’s modulus and Poisson’s Ratio. In this experiment, we set Young's modulus to $2\cdot10^5$ Pa and Poisson's ratio to 0.3. The friction coefficient was set to 0.8. 

To test each grasp candidate in the simulation, we first initialize the gripper to a predefined pose and set it fully open. Then, the gripper is slowly closed. Once contact between the finger and the object is detected, we start recording the contact forces, contact locations, and contact normals. Next, using the recorded data, we compute the approximated friction cones to continuously calculate the \gls{gws}, which is subsequently used to determine the quality metric. To enable a fair comparison between the quality metrics, we look at the calculated quality metrics at a certain time step where the squeezing force reaches the predefined desired force.

To evaluate whether a grasp is successful or not, we used the \textit{linear instability metric} from the shake task provided by DefGraspSim as ground truth because this metric has previously been shown to perform well in quantifying the quality of a grasp on deformable objects \cite{defggcnn}. The linear instability metric is defined as the average acceleration over different directions that would push the object out of the gripper. A higher metric is considered better as it indicates that a grasp can withstand higher accelerations.

We compared the performance of the proposed quality metric to two popular existing quality metrics: the volume metric \cite{volumeofgws}, and the Ferrari-Canny epsilon metric \cite{ferraricanny}. In total, we evaluated 600 grasps per metric.

Ideally, if a quality metric linearly increases with the linear instability metric, that quality metric is considered to better represent a good grasp. We can therefore quantify how good a quality metric is by calculating its monotonicity with the linear instability metric. The higher and closer to 1 the monotonicity, the better the quality metric.

\begin{figure}[t]
	\centering
    \def\svgwidth{\linewidth}
     {\color{black} \fontsize{8}{8}
    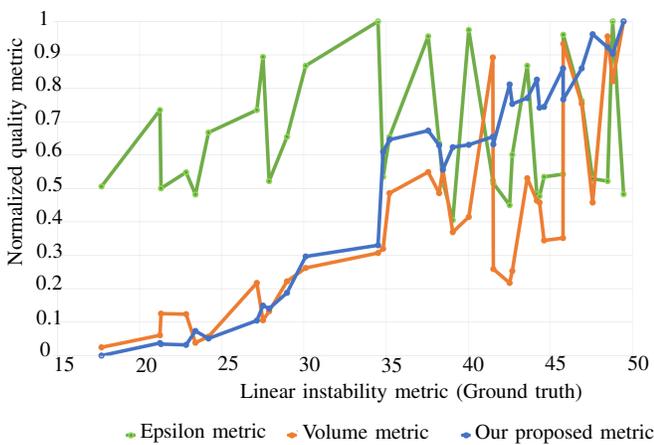}   
	\caption{The quality metric candidates versus the ground truth on the tall cylindrical object (object 1).}
	\label{fig:tall_cylinder}
\end{figure}

An example of the monotonicity for object 1 in \figref{fig:simobjects} is visualized in \figref{fig:tall_cylinder}. We can see that for this object, the proposed quality metric matches the ground truth much better than the epsilon and volume metric. Specifically, the monotonicity for our, the epsilon, and the volume metric are 99.9, 3.33, and 79.3, respectively. Among these three metrics, the epsilon metric performs worse due to strong oscillations. 

\begin{table}
    \centering
	\ra{1.3}\tbs{7}
	\caption{\label{tb:monotonicity} Average monotonicity of the quality metrics on all target objects. $\uparrow$: higher the better}
    \begin{tabular}{@{}lccc@{}}
        \toprule
        & \multicolumn{1}{l}{Epsilon metric} & \multicolumn{1}{l}{Volume metric}  & \multicolumn{1}{l}{Our metric}\\
        \midrule
        Monotonicity ($10^{-2}$) $\uparrow$ & 6   & 66  &\textbf{91} \\
        \bottomrule
    \end{tabular}
     \figvspace{}
\end{table}
\tabref{tb:monotonicity} presents the average monotonicity for all \glspl{gqm} over all objects and grasps. These results clearly show that the proposed quality metric achieves the highest average monotonicity and is significantly better than the other two metrics. Again, similar to the example above, the epsilon metric exhibits the worst performance. 

Interestingly, the average monotonicity of the volume metric compared to the example above is much worse: 66.3 compared to 79.3. This decline shows that the performance of the volume metric heavily depends on the geometry and shape of the deformable object. One reason for that is due to the known fact that the volume metric is incapable of indicating whether a particular
direction can only resist small disturbing forces. And as the object's geometric complexity increase, the gravity-induced deformations start to affect more on the grasp configuration which would confuse the volume metric.  Therefore, the performance of the volume metric drops when we start to introduce new and complex objects to the object set.

\begin{figure}[t]
	\centering
    \def\svgwidth{\linewidth}
     {\color{black} \fontsize{8}{8}
    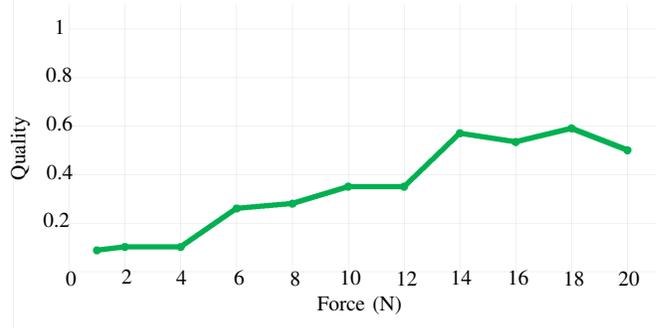}   
	\caption{Average quality metric versus the grasping force on all objects.}
	\label{fig:forcevsqall}
	\vspace{-1em}
\end{figure}

Another important aspect when grasping deformable objects is the grasping force. As shown in \figref{fig:title}, the higher the grasping force, the more contact points. To study the relationship between the grasp force and the grasp stability, and to show that our approach and \gls{gqm} can capture this relationship, we conducted an experiment where we linearly increased the grasping force from 1N to 20N and continuously evaluate the quality of the grasps over the different grasping forces. This experiment is evaluated on the same object set as in \figref{fig:simobjects}. The results, that are displayed in \figref{fig:forcevsqall}, demonstrate that the grasp quality increases linearly with the grasp force up to a specific force value. We hypothesize that this finding stems from the fact that at a certain grasp force the object reaches its maximum deformation, and from thereon the grasp quality cannot increase anymore. 
\begin{figure}[t]
	\centering
    \includegraphics[width=\linewidth]{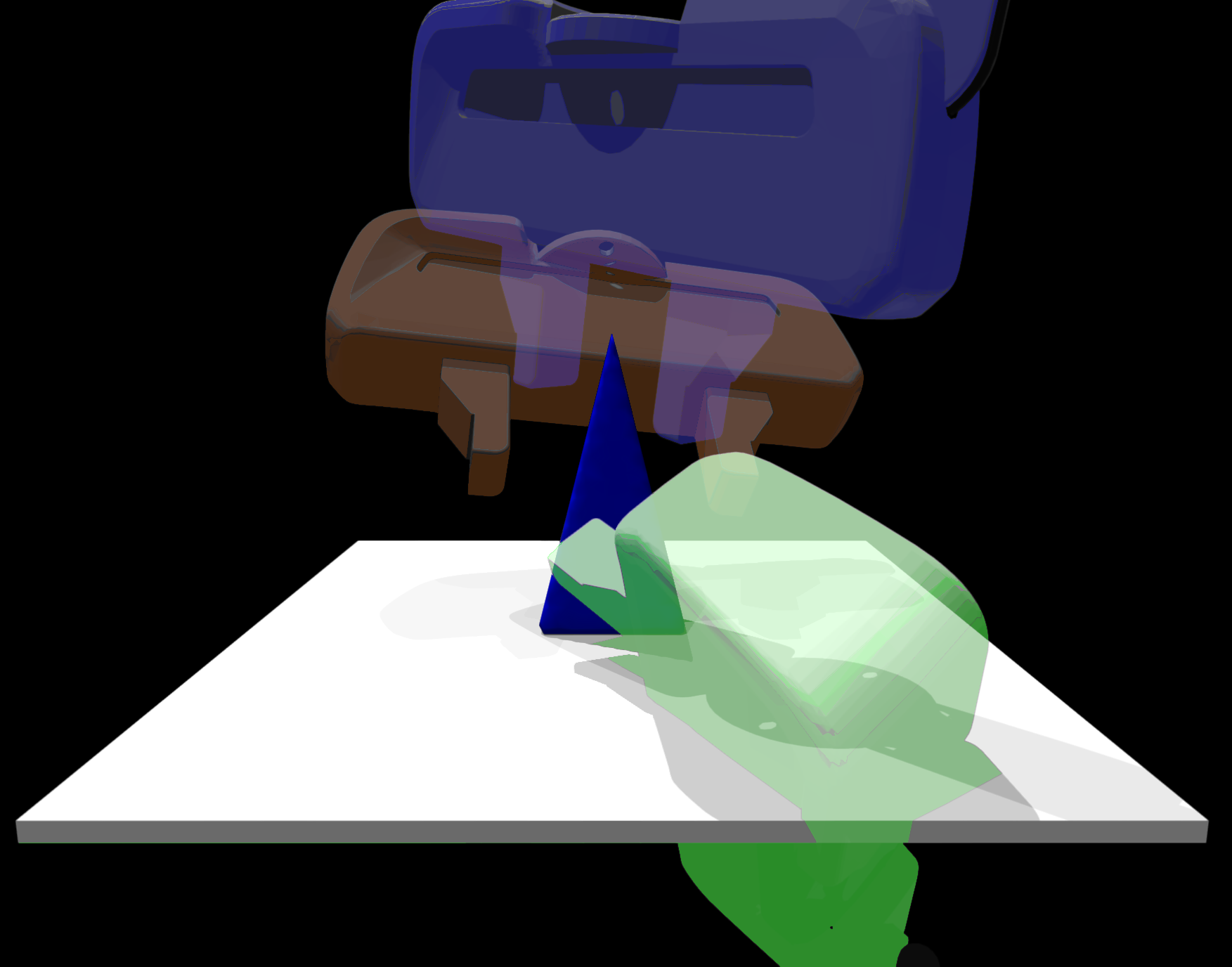} 
	\caption{There grasp candidates are colored as follows: grasp 2 (orange), grasp 5 (blue) and grasp 4 (green). Note that the collision between the gripper and the rectangular platform does not affect the result.}
	\label{fig:conegrasp}
	\vspace{-1em}
\end{figure}
\begin{figure}[t]
	\centering
    \def\svgwidth{\linewidth}
     {\color{black} \fontsize{8}{8}
    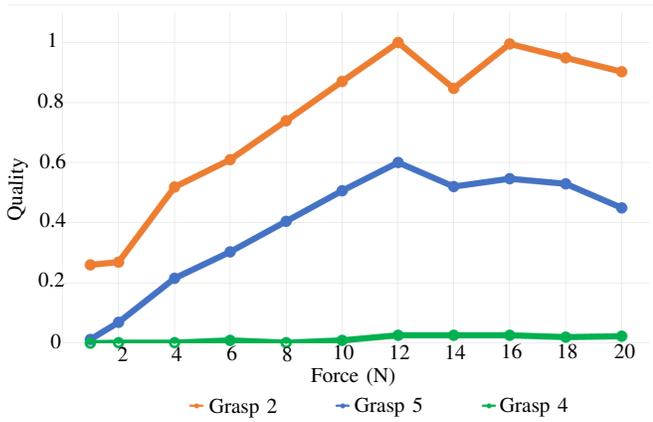}   
	\caption{Our proposed quality metric versus the grasping force on three grasp candidates.}
	\label{fig:forcevsq}
	\vspace{-1em}
\end{figure}

To study the effect the grasp force has on individual grasps, we evaluated, on one object, our quality metric over different grasping forces for the three grasp candidates shown in \figref{fig:conegrasp}. The result is presented in \figref{fig:forcevsq} and shows that grasp 4 fails to grasp the object regardless of the grasping force, while grasp 2 is better than grasp 5 in terms of grasp quality. Additionally, the quality of grasp 5 at 12N is approximately equal to that of grasp 2 at 6N. This finding indicates that we can plan similar quality grasps but with much less grasping force, which is extremely beneficial in the case of deformable objects that exhibit plastic deformation behavior where limiting the grasping force to a certain range is necessary to avoid permanent deformation. 

Finally, we compared the simulation time needed to evaluate the grasps using our metric and the ground truth. The result is shown in \tabref{tb:computetime}. As expected, the simulation time needed to compute the metrics increases as the complexity of the mesh increases, as indicated by the number of vertices. However, the results clearly show that the simulation time needed to compute our metric is significantly less than the ground truth, even when the number of vertices is very few. The lower simulation time using our metric is because computing the ground truth requires to simulate the whole \shake while computing our metric only requires to simulate the squeezing procedure.
\begin{table}
    \centering
	\ra{1.3}\tbs{4}
	\caption{\label{tb:computetime} Simulation time (seconds) required to compute the quality metrics on the target objects. $\downarrow$: lower the better}
    \begin{tabular}{@{}lcc@{}}
        \toprule
        Number of vertices & \multicolumn{1}{l}{Our metric} & \multicolumn{1}{l}{Linear instability metric (ground truth)}\\
        \midrule
                364             & 5    & 73         \\
                770             &   7.5  & 98        \\
                1198             & 7.9    & 199.27  \\
                2395            & 15.6   & 319.2       \\
                3182            & 35.8   & 380.7      \\
                5573            & 95.4   & 685.1    \\
        \bottomrule
    \end{tabular}
     \figvspace{}
\end{table}

\subsection{Grasp Planning in the Real World}
\begin{figure}
    \centering
    \def\svgwidth{\linewidth}
     {\color{black} \fontsize{15}{15}
    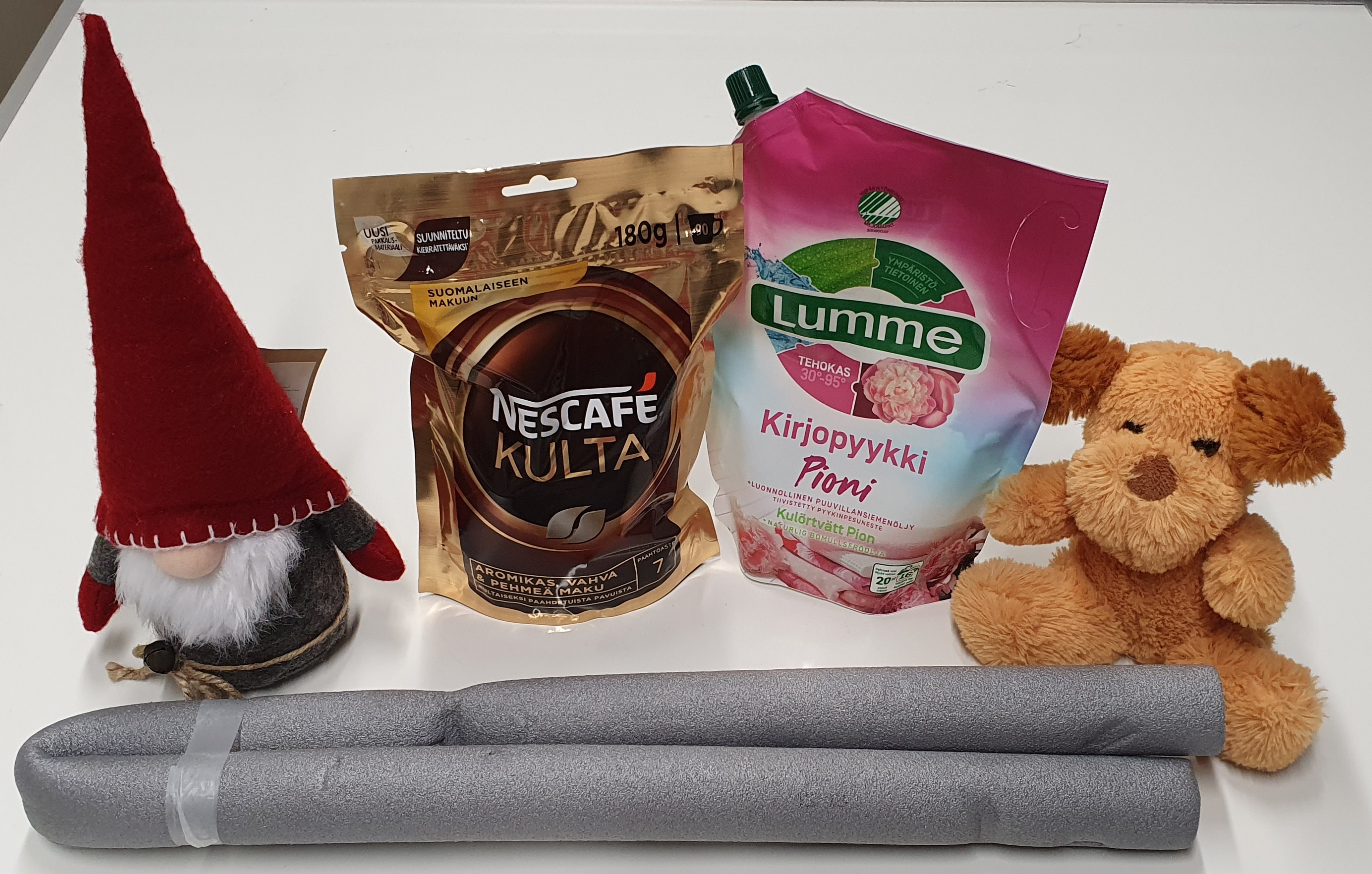}   
    \caption{The five objects used in the real-world experiment.}
    \label{fig:testobjreal}
\end{figure}

In the final experiment, we investigate if the high-quality grasps proposed by our quality metric transferred to real-world grasp success. To evaluate grasp success we used a \panda equipped with a parallel-jaw gripper. The robot was tasked to grasp the five objects shown in \figref{fig:testobjreal}. We chose these objects as they represent a high variation in size, shape, and mass distribution.

In order to calculate the quality metrics in Isaac Gym, we need a tetrahedral mesh of the object to grasp. As the objects were all unknown, we had to reconstruct them. To do the reconstruction, we first captured a multi-view point cloud of the object using an Intel RealSense D435 camera mounted to the robot's wrist. Then we converted the merged point cloud to a triangular mesh using Meshlab. Finally, we converted the triangular mesh into a tetrahedral mesh using fTetWild \cite{ftetwild}.

Using the reconstructed mesh, we then sampled 50 grasp candidates around the object and evaluated each one using our metric, the epsilon metric, and the volume metric. We chose the ten grasps with the highest quality for each of the qualities and executed them on the real robot. To evaluate if a grasp was successful, the robot moved to the planned grasp pose, closed its fingers, and then moved the arm back to the starting position. Once there, the robot executes a predefined trajectory that includes linear accelerating and reorienting the object. Finally, the robot rotates the gripper around the last joint and places the object at the goal position. A grasp was considered successful if the object was kept in the gripper during the whole procedure and unsuccessful if the object was dropped.


\begin{table}
    \centering
	\ra{1.3}\tbs{10}
	\caption{\label{tb:realexp} The average grasp success rate (\%) per quality metric on the target objects. $\uparrow$: higher the better}
    \begin{tabular}{@{}lccc@{}}
        \toprule
        Object & \multicolumn{1}{l}{Epsilon metric} & \multicolumn{1}{l}{Volume metric}  & \multicolumn{1}{l}{Our metric}\\
        \midrule
        1  & 50    & 50  & 70 \\
        2  & 60    & 80  & 80 \\
        3  & 60    & 60  & 80 \\
        4  & 80    & 100  &100 \\
        5 &  90    & 80  & 100\\
        \midrule
        All  $\uparrow$ & 68    & 74  &\textbf{86} \\
        \bottomrule
    \end{tabular}
     \figvspace{}
\end{table}

The experimental results are presented in \tabref{tb:realexp}. The results suggest that our proposed quality is more suitable for deformable objects grasping than the classical wrench-based quality metrics. Specifically, grasps ranked by the proposed metric achieve a 12\% higher grasp success rate than those ranked according to the volume metric and an 18\% higher success rate than those ranked by the epsilon metric. 

If we focus on the grasp success rate on each object separately, the result shows that all three quality metrics perform well on light objects such as object 4 and object 5. However, when it comes to objects with non-uniform mass distribution, the performance of our proposed quality metric surpasses that of the other two. The main reason for the poorly recognized grasps by the other quality metrics on these objects was that the robot often dropped the object once it started to accelerate or reorient the object due to the deformation caused by gravity. Our quality metric avoided such grasps as it was specifically formulated to account for the effect of unknown gravity directions, which resulted in gravity-resistant grasps.

%% file: figures_tex/simobjects.pdf_tex
\begingroup%
  \makeatletter%
  \providecommand\color[2][]{%
    \errmessage{(Inkscape) Color is used for the text in Inkscape, but the package 'color.sty' is not loaded}%
    \renewcommand\color[2][]{}%
  }%
  \providecommand\transparent[1]{%
    \errmessage{(Inkscape) Transparency is used (non-zero) for the text in Inkscape, but the package 'transparent.sty' is not loaded}%
    \renewcommand\transparent[1]{}%
  }%
  \providecommand\rotatebox[2]{#2}%
  \newcommand*\fsize{\dimexpr\f@size pt\relax}%
  \newcommand*\lineheight[1]{\fontsize{\fsize}{#1\fsize}\selectfont}%
  \ifx\svgwidth\undefined%
    \setlength{\unitlength}{1098.0000173bp}%
    \ifx\svgscale\undefined%
      \relax%
    \else%
      \setlength{\unitlength}{\unitlength * \real{\svgscale}}%
    \fi%
  \else%
    \setlength{\unitlength}{\svgwidth}%
  \fi%
  \global\let\svgwidth\undefined%
  \global\let\svgscale\undefined%
  \makeatother%
  \begin{picture}(1,0.52254098)%
    \lineheight{1}%
    \setlength\tabcolsep{0pt}%
    \put(0,0){\includegraphics[width=\unitlength,page=1]{simobjects.pdf}}%
    \put(0.12623377,0.4677007){\makebox(0,0)[lt]{\lineheight{1.25}\smash{\begin{tabular}[t]{l}1\end{tabular}}}}%
    \put(0.26895667,0.46368612){\makebox(0,0)[lt]{\lineheight{1.25}\smash{\begin{tabular}[t]{l}2\end{tabular}}}}%
    \put(0.42558346,0.46107769){\makebox(0,0)[lt]{\lineheight{1.25}\smash{\begin{tabular}[t]{l}3\end{tabular}}}}%
    \put(0.56857468,0.46048453){\makebox(0,0)[lt]{\lineheight{1.25}\smash{\begin{tabular}[t]{l}4\end{tabular}}}}%
    \put(0.72730678,0.45216476){\makebox(0,0)[lt]{\lineheight{1.25}\smash{\begin{tabular}[t]{l}5\end{tabular}}}}%
    \put(0.06323386,0.18244894){\makebox(0,0)[lt]{\lineheight{1.25}\smash{\begin{tabular}[t]{l}6\end{tabular}}}}%
    \put(0.21998258,0.17380924){\makebox(0,0)[lt]{\lineheight{1.25}\smash{\begin{tabular}[t]{l}7\end{tabular}}}}%
    \put(0.35284875,0.17778172){\makebox(0,0)[lt]{\lineheight{1.25}\smash{\begin{tabular}[t]{l}8\end{tabular}}}}%
    \put(0.54516785,0.17851147){\makebox(0,0)[lt]{\lineheight{1.25}\smash{\begin{tabular}[t]{l}9\end{tabular}}}}%
    \put(0.72891183,0.18316608){\makebox(0,0)[lt]{\lineheight{1.25}\smash{\begin{tabular}[t]{l}10\end{tabular}}}}%
  \end{picture}%
\endgroup%

%% file: figures_tex/tall_cylinder.pdf_tex
\begingroup%
  \makeatletter%
  \providecommand\color[2][]{%
    \errmessage{(Inkscape) Color is used for the text in Inkscape, but the package 'color.sty' is not loaded}%
    \renewcommand\color[2][]{}%
  }%
  \providecommand\transparent[1]{%
    \errmessage{(Inkscape) Transparency is used (non-zero) for the text in Inkscape, but the package 'transparent.sty' is not loaded}%
    \renewcommand\transparent[1]{}%
  }%
  \providecommand\rotatebox[2]{#2}%
  \newcommand*\fsize{\dimexpr\f@size pt\relax}%
  \newcommand*\lineheight[1]{\fontsize{\fsize}{#1\fsize}\selectfont}%
  \ifx\svgwidth\undefined%
    \setlength{\unitlength}{1828.82691391bp}%
    \ifx\svgscale\undefined%
      \relax%
    \else%
      \setlength{\unitlength}{\unitlength * \real{\svgscale}}%
    \fi%
  \else%
    \setlength{\unitlength}{\svgwidth}%
  \fi%
  \global\let\svgwidth\undefined%
  \global\let\svgscale\undefined%
  \makeatother%
  \begin{picture}(1,0.69770304)%
    \lineheight{1}%
    \setlength\tabcolsep{0pt}%
    \put(0,0){\includegraphics[width=\unitlength,page=1]{tall_cylinder.pdf}}%
    \put(0.05027857,0.14687959){\makebox(0,0)[lt]{\lineheight{1.25}\smash{\begin{tabular}[t]{l}0\end{tabular}}}}%
    \put(0.04477645,0.19639869){\makebox(0,0)[lt]{\lineheight{1.25}\smash{\begin{tabular}[t]{l}0.1\end{tabular}}}}%
    \put(0.04416512,0.24897446){\makebox(0,0)[lt]{\lineheight{1.25}\smash{\begin{tabular}[t]{l}0.2\end{tabular}}}}%
    \put(0.0453878,0.29849356){\makebox(0,0)[lt]{\lineheight{1.25}\smash{\begin{tabular}[t]{l}0.3\end{tabular}}}}%
    \put(0.04477647,0.35045806){\makebox(0,0)[lt]{\lineheight{1.25}\smash{\begin{tabular}[t]{l}0.4\end{tabular}}}}%
    \put(0.04538782,0.40181119){\makebox(0,0)[lt]{\lineheight{1.25}\smash{\begin{tabular}[t]{l}0.5\end{tabular}}}}%
    \put(0.04355375,0.45438701){\makebox(0,0)[lt]{\lineheight{1.25}\smash{\begin{tabular}[t]{l}0.6\end{tabular}}}}%
    \put(0.04294242,0.50390611){\makebox(0,0)[lt]{\lineheight{1.25}\smash{\begin{tabular}[t]{l}0.7\end{tabular}}}}%
    \put(0.04355375,0.55587058){\makebox(0,0)[lt]{\lineheight{1.25}\smash{\begin{tabular}[t]{l}0.8\end{tabular}}}}%
    \put(0.04171972,0.60661236){\makebox(0,0)[lt]{\lineheight{1.25}\smash{\begin{tabular}[t]{l}0.9\end{tabular}}}}%
    \put(0.04844452,0.65735416){\makebox(0,0)[lt]{\lineheight{1.25}\smash{\begin{tabular}[t]{l}1\end{tabular}}}}%
    \put(0.07131875,0.12361792){\makebox(0,0)[lt]{\lineheight{1.25}\smash{\begin{tabular}[t]{l}15\end{tabular}}}}%
    \put(0.19807637,0.1227977){\makebox(0,0)[lt]{\lineheight{1.25}\smash{\begin{tabular}[t]{l}20\end{tabular}}}}%
    \put(0.32401381,0.12585443){\makebox(0,0)[lt]{\lineheight{1.25}\smash{\begin{tabular}[t]{l}25\end{tabular}}}}%
    \put(0.44933988,0.12482534){\makebox(0,0)[lt]{\lineheight{1.25}\smash{\begin{tabular}[t]{l}30\end{tabular}}}}%
    \put(0.57588868,0.12279766){\makebox(0,0)[lt]{\lineheight{1.25}\smash{\begin{tabular}[t]{l}35\end{tabular}}}}%
    \put(0.70121479,0.12342426){\makebox(0,0)[lt]{\lineheight{1.25}\smash{\begin{tabular}[t]{l}40\end{tabular}}}}%
    \put(0.82797243,0.12567606){\makebox(0,0)[lt]{\lineheight{1.25}\smash{\begin{tabular}[t]{l}45\end{tabular}}}}%
    \put(0.95492371,0.12586971){\makebox(0,0)[lt]{\lineheight{1.25}\smash{\begin{tabular}[t]{l}50\end{tabular}}}}%
    \put(0.35854119,0.08216375){\makebox(0,0)[lt]{\lineheight{1.25}\smash{\begin{tabular}[t]{l}Linear instability metric (Ground truth)\end{tabular}}}}%
    \put(0.02269803,0.28709675){\rotatebox{90}{\makebox(0,0)[lt]{\lineheight{1.25}\smash{\begin{tabular}[t]{l}Normalized quality metric\end{tabular}}}}}%
    \put(0.20480117,0.02174715){\makebox(0,0)[lt]{\lineheight{1.25}\smash{\begin{tabular}[t]{l}Epsilon metric\end{tabular}}}}%
    \put(0.45423069,0.01991303){\makebox(0,0)[lt]{\lineheight{1.25}\smash{\begin{tabular}[t]{l}Volume metric\end{tabular}}}}%
    \put(0.7201665,0.01989785){\makebox(0,0)[lt]{\lineheight{1.25}\smash{\begin{tabular}[t]{l}Our proposed metric\end{tabular}}}}%
  \end{picture}%
\endgroup%

%% file: figures_tex/forcevsqall.pdf_tex
\begingroup%
  \makeatletter%
  \providecommand\color[2][]{%
    \errmessage{(Inkscape) Color is used for the text in Inkscape, but the package 'color.sty' is not loaded}%
    \renewcommand\color[2][]{}%
  }%
  \providecommand\transparent[1]{%
    \errmessage{(Inkscape) Transparency is used (non-zero) for the text in Inkscape, but the package 'transparent.sty' is not loaded}%
    \renewcommand\transparent[1]{}%
  }%
  \providecommand\rotatebox[2]{#2}%
  \newcommand*\fsize{\dimexpr\f@size pt\relax}%
  \newcommand*\lineheight[1]{\fontsize{\fsize}{#1\fsize}\selectfont}%
  \ifx\svgwidth\undefined%
    \setlength{\unitlength}{1307.99995963bp}%
    \ifx\svgscale\undefined%
      \relax%
    \else%
      \setlength{\unitlength}{\unitlength * \real{\svgscale}}%
    \fi%
  \else%
    \setlength{\unitlength}{\svgwidth}%
  \fi%
  \global\let\svgwidth\undefined%
  \global\let\svgscale\undefined%
  \makeatother%
  \begin{picture}(1,0.52006882)%
    \lineheight{1}%
    \setlength\tabcolsep{0pt}%
    \put(0,0){\includegraphics[width=\unitlength,page=1]{forcevsqall.pdf}}%
    \put(0.08089733,0.07846087){\makebox(0,0)[lt]{\lineheight{1.25}\smash{\begin{tabular}[t]{l}0\end{tabular}}}}%
    \put(0.16696268,0.081018){\makebox(0,0)[lt]{\lineheight{1.25}\smash{\begin{tabular}[t]{l}2\end{tabular}}}}%
    \put(0.25138677,0.07940159){\makebox(0,0)[lt]{\lineheight{1.25}\smash{\begin{tabular}[t]{l}4\end{tabular}}}}%
    \put(0.33991785,0.07916496){\makebox(0,0)[lt]{\lineheight{1.25}\smash{\begin{tabular}[t]{l}6\end{tabular}}}}%
    \put(0.4260779,0.07849602){\makebox(0,0)[lt]{\lineheight{1.25}\smash{\begin{tabular}[t]{l}8\end{tabular}}}}%
    \put(0.50307235,0.07916496){\makebox(0,0)[lt]{\lineheight{1.25}\smash{\begin{tabular}[t]{l}10\end{tabular}}}}%
    \put(0.58992777,0.07867363){\makebox(0,0)[lt]{\lineheight{1.25}\smash{\begin{tabular}[t]{l}12\end{tabular}}}}%
    \put(0.67347591,0.08002837){\makebox(0,0)[lt]{\lineheight{1.25}\smash{\begin{tabular}[t]{l}14\end{tabular}}}}%
    \put(0.7614049,0.07910447){\makebox(0,0)[lt]{\lineheight{1.25}\smash{\begin{tabular}[t]{l}16\end{tabular}}}}%
    \put(0.84711937,0.07868335){\makebox(0,0)[lt]{\lineheight{1.25}\smash{\begin{tabular}[t]{l}18\end{tabular}}}}%
    \put(0.93178406,0.07872092){\makebox(0,0)[lt]{\lineheight{1.25}\smash{\begin{tabular}[t]{l}20\end{tabular}}}}%
    \put(0.04634999,0.16894451){\makebox(0,0)[lt]{\lineheight{1.25}\smash{\begin{tabular}[t]{l}0.2\end{tabular}}}}%
    \put(0.05157187,0.24264526){\makebox(0,0)[lt]{\lineheight{1.25}\smash{\begin{tabular}[t]{l}0.4\end{tabular}}}}%
    \put(0.05105093,0.31786564){\makebox(0,0)[lt]{\lineheight{1.25}\smash{\begin{tabular}[t]{l}0.6\end{tabular}}}}%
    \put(0.05071433,0.39260431){\makebox(0,0)[lt]{\lineheight{1.25}\smash{\begin{tabular}[t]{l}0.8\end{tabular}}}}%
    \put(0.06359457,0.46615438){\makebox(0,0)[lt]{\lineheight{1.25}\smash{\begin{tabular}[t]{l}1\end{tabular}}}}%
    \put(0.46841253,0.03971269){\makebox(0,0)[lt]{\lineheight{1.25}\smash{\begin{tabular}[t]{l}Force (N)\end{tabular}}}}%
    \put(0.01905118,0.24126727){\rotatebox{90}{\makebox(0,0)[lt]{\lineheight{1.25}\smash{\begin{tabular}[t]{l}Quality\end{tabular}}}}}%
  \end{picture}%
\endgroup%

%% file: figures_tex/forcevsq.pdf_tex
\begingroup%
  \makeatletter%
  \providecommand\color[2][]{%
    \errmessage{(Inkscape) Color is used for the text in Inkscape, but the package 'color.sty' is not loaded}%
    \renewcommand\color[2][]{}%
  }%
  \providecommand\transparent[1]{%
    \errmessage{(Inkscape) Transparency is used (non-zero) for the text in Inkscape, but the package 'transparent.sty' is not loaded}%
    \renewcommand\transparent[1]{}%
  }%
  \providecommand\rotatebox[2]{#2}%
  \newcommand*\fsize{\dimexpr\f@size pt\relax}%
  \newcommand*\lineheight[1]{\fontsize{\fsize}{#1\fsize}\selectfont}%
  \ifx\svgwidth\undefined%
    \setlength{\unitlength}{1054.58386519bp}%
    \ifx\svgscale\undefined%
      \relax%
    \else%
      \setlength{\unitlength}{\unitlength * \real{\svgscale}}%
    \fi%
  \else%
    \setlength{\unitlength}{\svgwidth}%
  \fi%
  \global\let\svgwidth\undefined%
  \global\let\svgscale\undefined%
  \makeatother%
  \begin{picture}(1,0.6458037)%
    \lineheight{1}%
    \setlength\tabcolsep{0pt}%
    \put(0,0){\includegraphics[width=\unitlength,page=1]{forcevsq.pdf}}%
    \put(0.31063241,0.01711763){\makebox(0,0)[lt]{\lineheight{1.25}\smash{\begin{tabular}[t]{l}Grasp 2\end{tabular}}}}%
    \put(0.53433014,0.01817785){\makebox(0,0)[lt]{\lineheight{1.25}\smash{\begin{tabular}[t]{l}Grasp 5\end{tabular}}}}%
    \put(0.75802777,0.01923808){\makebox(0,0)[lt]{\lineheight{1.25}\smash{\begin{tabular}[t]{l}Grasp 4\end{tabular}}}}%
    \put(0.06679126,0.11677444){\makebox(0,0)[lt]{\lineheight{1.25}\smash{\begin{tabular}[t]{l}0\end{tabular}}}}%
    \put(0.16962862,0.09663106){\makebox(0,0)[lt]{\lineheight{1.25}\smash{\begin{tabular}[t]{l}2\end{tabular}}}}%
    \put(0.25302055,0.09732904){\makebox(0,0)[lt]{\lineheight{1.25}\smash{\begin{tabular}[t]{l}4\end{tabular}}}}%
    \put(0.33783486,0.09763862){\makebox(0,0)[lt]{\lineheight{1.25}\smash{\begin{tabular}[t]{l}6\end{tabular}}}}%
    \put(0.42301134,0.09663102){\makebox(0,0)[lt]{\lineheight{1.25}\smash{\begin{tabular}[t]{l}8\end{tabular}}}}%
    \put(0.50141184,0.09975899){\makebox(0,0)[lt]{\lineheight{1.25}\smash{\begin{tabular}[t]{l}10\end{tabular}}}}%
    \put(0.58733908,0.09939679){\makebox(0,0)[lt]{\lineheight{1.25}\smash{\begin{tabular}[t]{l}12\end{tabular}}}}%
    \put(0.67321354,0.09981161){\makebox(0,0)[lt]{\lineheight{1.25}\smash{\begin{tabular}[t]{l}14\end{tabular}}}}%
    \put(0.76014826,0.10193194){\makebox(0,0)[lt]{\lineheight{1.25}\smash{\begin{tabular}[t]{l}16\end{tabular}}}}%
    \put(0.84602268,0.10193198){\makebox(0,0)[lt]{\lineheight{1.25}\smash{\begin{tabular}[t]{l}18\end{tabular}}}}%
    \put(0.93401759,0.10193194){\makebox(0,0)[lt]{\lineheight{1.25}\smash{\begin{tabular}[t]{l}20\end{tabular}}}}%
    \put(0.04770804,0.21113034){\makebox(0,0)[lt]{\lineheight{1.25}\smash{\begin{tabular}[t]{l}0.2\end{tabular}}}}%
    \put(0.04982841,0.30336594){\makebox(0,0)[lt]{\lineheight{1.25}\smash{\begin{tabular}[t]{l}0.4\end{tabular}}}}%
    \put(0.04982841,0.39454132){\makebox(0,0)[lt]{\lineheight{1.25}\smash{\begin{tabular}[t]{l}0.6\end{tabular}}}}%
    \put(0.04664787,0.48783705){\makebox(0,0)[lt]{\lineheight{1.25}\smash{\begin{tabular}[t]{l}0.8\end{tabular}}}}%
    \put(0.06149033,0.58113279){\makebox(0,0)[lt]{\lineheight{1.25}\smash{\begin{tabular}[t]{l}1\end{tabular}}}}%
    \put(0.46753885,0.06513526){\makebox(0,0)[lt]{\lineheight{1.25}\smash{\begin{tabular}[t]{l}Force (N)\end{tabular}}}}%
    \put(0.02170082,0.31276437){\rotatebox{87.81687}{\makebox(0,0)[lt]{\lineheight{1.25}\smash{\begin{tabular}[t]{l}Quality\end{tabular}}}}}%
  \end{picture}%
\endgroup%

%% file: figures_tex/realobjects.pdf_tex
\begingroup%
  \makeatletter%
  \providecommand\color[2][]{%
    \errmessage{(Inkscape) Color is used for the text in Inkscape, but the package 'color.sty' is not loaded}%
    \renewcommand\color[2][]{}%
  }%
  \providecommand\transparent[1]{%
    \errmessage{(Inkscape) Transparency is used (non-zero) for the text in Inkscape, but the package 'transparent.sty' is not loaded}%
    \renewcommand\transparent[1]{}%
  }%
  \providecommand\rotatebox[2]{#2}%
  \newcommand*\fsize{\dimexpr\f@size pt\relax}%
  \newcommand*\lineheight[1]{\fontsize{\fsize}{#1\fsize}\selectfont}%
  \ifx\svgwidth\undefined%
    \setlength{\unitlength}{2768.00006104bp}%
    \ifx\svgscale\undefined%
      \relax%
    \else%
      \setlength{\unitlength}{\unitlength * \real{\svgscale}}%
    \fi%
  \else%
    \setlength{\unitlength}{\svgwidth}%
  \fi%
  \global\let\svgwidth\undefined%
  \global\let\svgscale\undefined%
  \makeatother%
  \begin{picture}(1,0.63728322)%
    \lineheight{1}%
    \setlength\tabcolsep{0pt}%
    \put(0,0){\includegraphics[width=\unitlength,page=1]{realobjects.pdf}}%
    \put(0.11761749,0.53373177){\makebox(0,0)[lt]{\lineheight{1.25}\smash{\begin{tabular}[t]{l}1\end{tabular}}}}%
    \put(0.36755465,0.53545173){\makebox(0,0)[lt]{\lineheight{1.25}\smash{\begin{tabular}[t]{l}2\end{tabular}}}}%
    \put(0.68028245,0.54735014){\makebox(0,0)[lt]{\lineheight{1.25}\smash{\begin{tabular}[t]{l}3\end{tabular}}}}%
    \put(0.84697788,0.40544106){\makebox(0,0)[lt]{\lineheight{1.25}\smash{\begin{tabular}[t]{l}4\end{tabular}}}}%
    \put(0.51457653,0.15889671){\makebox(0,0)[lt]{\lineheight{1.25}\smash{\begin{tabular}[t]{l}5\end{tabular}}}}%
  \end{picture}%
\endgroup%

%% file: sections/conclusion.tex
\section{Conclusions and future work}
\label{sec:conclusions}
Although research on manipulating 3D deformable objects is starting to receive more and more attention, grasp evaluation on deformable objects is still under-explored, especially if we compare to grasp evaluation on rigid objects. In this paper, we took a step to close this gap by introducing an approach to simulate the interaction between the robot and the object and predict their contact dynamics, together with a quality metric to evaluate grasp candidates on deformable objects. The key idea was to utilize a \gls{fem}-based soft-body physics simulator to continuously compute and capture changes in shape and contact geometry between the deformable object and the gripper during the grasp. In addition, we also propose a grasp quality metric that targets deformable objects by combining the grasp wrench resistance and the object deformation. 

The experimental results from simulation demonstrate that the proposed approach is more suitable for evaluating the quality of a grasp on a deformable object than classical wrench-based quality metrics. Similar observations were also seen in the physical grasping experiment. Our quality metric is also up to fifteen times faster to calculate than the \shake, which, to date, is one of the most reliable approaches to quantify a grasp on a deformable object.

Because of the apparent correlation between the proposed metric and successful grasps and the speed to evaluate the quality metric, this work opens up the ability to generate large-scale datasets of millions of grasps on deformable objects. This, in turn, would allow training deformation-aware grasp samplers such as the one in \cite{defggcnn} on a much richer dataset which would hopefully improve grasp success rates across many more object stiffnesses.